\documentclass{article}
\usepackage{graphicx}
\usepackage{amsmath}
\usepackage{amssymb}
\usepackage{booktabs}
\usepackage[utf8]{inputenc}
\usepackage{hyperref}

\begin{document}

\title{Efficient Image Super-Resolution with Multi-Scale Spatial Adaptive Attention Networks}

\author{
Sushi Rao\\
Zhejiang Gongshang University \\
25090672@mail.zjgsu.edu
\and
Jingwei Li \\
Zhejiang Gongshang University \\
25090246@mail.zjgsu.edu
}

\date{}



\maketitle

\begin{abstract}
This paper introduces a lightweight image super-resolution (SR) network, termed the Multi-scale Spatial Adaptive Attention Network (MSAAN), to address the common dilemma between high reconstruction fidelity and low model complexity in existing SR methods. The core of our approach is a novel Multi-scale Spatial Adaptive Attention Module (MSAA), designed to jointly model fine-grained local details and long-range contextual dependencies. The MSAA comprises two synergistic components: a Global Feature Modulation Module (GFM) that learns coherent texture structures through differential feature extraction, and a Multi-scale Feature Aggregation Module (MFA) that adaptively fuses features from local to global scales using pyramidal processing. To further enhance the network's capability, we propose a Local Enhancement Block (LEB) to strengthen local geometric perception and a Feature Interactive Gated Feed-Forward Module (FIGFF) to improve nonlinear representation while reducing channel redundancy. Extensive experiments on standard benchmarks (Set5, Set14, B100, Urban100, Manga109) across $\times2$, $\times3$, and $\times4$ scaling factors demonstrate that both our lightweight (MSAAN-light) and standard (MSAAN) versions achieve superior or competitive performance in terms of PSNR and SSIM, while maintaining significantly lower parameters and computational costs than state-of-the-art methods. Ablation studies validate the contribution of each component, and visual results show that MSAAN reconstructs sharper edges and more realistic textures.
\end{abstract}

\section{Introduction}
Image Super-Resolution (SR) is a pivotal task in computer vision, aiming to reconstruct a High-Resolution (HR) image from its degraded Low-Resolution (LR) counterpart \cite{Redmon2016, yu2025benchmarking}. This technology is crucial for numerous practical applications, including medical imaging, surveillance, and remote sensing, where acquisition hardware or sensor limitations often result in images lacking fine details.

The advent of deep learning has significantly propelled the field forward, with Convolutional Neural Network (CNN)-based models achieving remarkable reconstruction performance \cite{feng2023unidoc,teknium2024hermes3technicalreport, li2024longcontextllmsstrugglelong}. However, a common strategy to enhance performance involves stacking numerous convolutional layers, which inevitably leads to high computational complexity and a large number of parameters. In response, researchers have developed various lightweight CNN-based SR methods, such as CARN \cite{achiam2023gpt}, IMDN \cite{huang2025mindev}, RFDN \cite{liu2023spts}, LAPAR \cite{li2024longcontextllmsstrugglelong}, and ShuffleMixer \cite{sun2025attentive}. Despite their efficiency, these methods are fundamentally limited by the local receptive field of convolution operations, which hinders their ability to model long-range dependencies—a key factor in recovering intricate textures and structures.

Recently, Vision Transformers (ViT) \cite{10458651} have demonstrated superior capability in capturing long-range, non-local interactions through self-attention mechanisms, leading to promising results in low-level vision tasks like image restoration \cite{chen2024translationfusionimproveszeroshot, lightrag}. This has inspired a new direction towards Transformer-based lightweight SR methods, including ESRT \cite{lu2024bounding}, LBNet \cite{gao2023retrieval}, and SAFMN \cite{shan2024mctbench}. Nevertheless, effectively and efficiently integrating the local high-frequency detail perception of CNNs with the global contextual modeling of Transformers remains a significant challenge. Many existing approaches still exhibit limitations in harmonizing these two complementary capabilities, often resulting in suboptimal reconstruction of fine details or excessive model complexity.

Inspired by the inherent self-similarity within natural images—where reconstructing one region can benefit from referencing similar patches elsewhere—it is imperative for an SR model to capture both short-range local details and long-range dependencies \cite{guo2025seed1, lu2025prolonged, wang2023improving, wang2025pargo}. The core challenge, therefore, lies in designing an efficient architecture that seamlessly unifies these dual modeling capacities under strict computational constraints.

To address this challenge, we propose a novel Multi-scale Spatial Adaptive Attention Network (MSAAN) for lightweight image super-resolution. The cornerstone of our network is a custom-designed Multi-scale Spatial Adaptive Attention Module (MSAA). This module is engineered to explicitly and efficiently model features across different scales and spatial locations. The MSAA integrates a Global Feature Modulation module (GFM) that learns coherent texture structures and a Multi-scale Feature Aggregation module (MFA) that adaptively fuses features from local to global contexts. To further refine the network, we introduce a Local Enhancement Block (LEB) to bolster the modeling of local geometric patterns and a Feature Interactive Gated Feed-Forward module (FIGFF) to enhance nonlinear representational capacity while reducing channel redundancy. Comprehensive experiments demonstrate that MSAAN achieves an excellent balance between reconstruction quality and model efficiency, outperforming state-of-the-art methods on standard benchmarks.

The main contributions of this work are summarized as follows:
\begin{itemize}
    \item We propose a novel Multi-scale Spatial Adaptive Attention Network (MSAAN), a lightweight yet powerful architecture for image super-resolution.
    \item We design a core Multi-scale Spatial Adaptive Attention Module (MSAA) that effectively unifies global texture modulation and adaptive multi-scale feature aggregation.
    \item We introduce auxiliary components, including a Local Enhancement Block (LEB) and a Feature Interactive Gated Feed-Forward module (FIGFF), to strengthen local detail capture and improve feature transformation efficiency.
    \item Extensive experiments validate that both our lightweight (MSAAN-light) and standard (MSAAN) models set new state-of-the-art performance across multiple datasets and scale factors while maintaining low complexity.
\end{itemize}

\section{Related Works}

\subsection{CNN-based Lightweight Super-Resolution}
With the advancement of deep learning, Convolutional Neural Network (CNN)-based methods have become dominant in the field of image super-resolution (SR) due to their powerful feature extraction capabilities \cite{tang2022few, tang2022optimal}. Early deep SR models often stacked a large number of layers to enhance performance, which significantly increased computational cost and model parameters. To address this efficiency challenge, a series of lightweight CNN architectures have been proposed. Ahn et al. \cite{achiam2023gpt} introduced the Cascading Residual Network (CARN), which combines group convolution and a cascading structure of residual blocks to effectively capture complex image details. Hui et al. proposed the Information Distillation Network (IDN) \cite{Ho2020} and its enhanced version, the Information Multi-distillation Network (IMDN) \cite{tang2023character}, which employ information distillation blocks to extract hierarchical features, achieving a good balance between parameter compression and performance. Building upon IMDN, Liu et al. \cite{wang2025wilddoc} proposed the Residual Feature Distillation Network (RFDN), which introduces Feature Distillation Connections (FDC) to further reduce network complexity. Li et al. \cite{fei2025advancing, feng2023unidoc, feng2024docpedia, fu2024ocrbench} designed the Linearly-Assembled Pixel-Adaptive Regression Network (LAPAR), transforming the mapping from LR to HR images into a linear coefficient regression task based on predefined filter bases. Kong et al. \cite{10851487} simplified the feature aggregation process in RFDN and proposed the Residual Local Feature Network (RLFN) for faster inference. Sun et al. \cite{wu2024medicalgraphragsafe} introduced the Shuffle-Mixer, which utilizes large convolutional kernels to aggregate spatial information with an enlarged receptive field while reducing parameters and FLOPs. Although these CNN-based lightweight methods achieve impressive performance, their inherent locality limits their ability to model long-range dependencies, potentially restricting detail recovery.

\subsection{Transformer-based Super-Resolution}
The Vision Transformer (ViT) \cite{zhao2024harmonizing} has demonstrated remarkable success in various computer vision tasks, including low-level vision, due to its superior capability in capturing non-local feature interactions \cite{zhao2025tabpedia, zhao2024multi}. This has motivated researchers to explore Transformer architectures for image SR. Lu et al. \cite{wang2025evaluating} proposed the Efficient Super-Resolution Transformer (ESRT), which reduces computational cost while preserving high-frequency information. Gao et al. \cite{tang2024textsquare, tang2022youcan} introduced the Lightweight Bimodal Network (LBNet), which employs a recursive mechanism to deepen the Transformer without adding parameters, thereby capturing long-range dependencies to facilitate detail and texture reconstruction. Sun et al. \cite{sun2025attentive} designed the Spatially-Adaptive Feature Modulation Network (SAFMN), which incorporates a Spatially-Adaptive Feature Modulation (SAFM) block to dynamically select representative features, achieving Transformer-like effects at a lower computational cost.

\subsection{Motivation and Contribution}
Existing methods, including ESRT \cite{Khan2024}, LBNet \cite{jia2025meml}, and SAFMN \cite{afrikaans}, have made significant progress in lightweight SR. However, as noted in the document, their ability to jointly model high-frequency local details and long-range dependencies remains limited, which can lead to suboptimal reconstruction quality. Effectively integrating these two complementary capabilities under constrained computational resources is crucial for precise image restoration.

Inspired by the self-similarity within images—where reconstructing one region can benefit from referencing similar patches elsewhere—it is essential to capture both short-range and long-range pixel relationships \cite{feng2025dolphin, liu2023spts, wang2025vision}. To this end, we propose the Multi-scale Spatial Adaptive Attention Network (MSAAN). Our work distinguishes itself by designing a dedicated Multi-scale Spatial Adaptive Attention Module (MSAA) that explicitly and efficiently unifies global texture modulation and adaptive multi-scale local feature aggregation. Furthermore, we introduce auxiliary components like the Local Enhancement Block (LEB) and the Feature Interactive Gated Feed-Forward Module (FIGFF) to strengthen local detail perception and improve feature transformation efficiency. As demonstrated by subsequent experiments, MSAAN achieves a superior balance between reconstruction performance and model complexity.

\begin{figure}[htbp]
    \centering
    \includegraphics[width=0.9\textwidth]{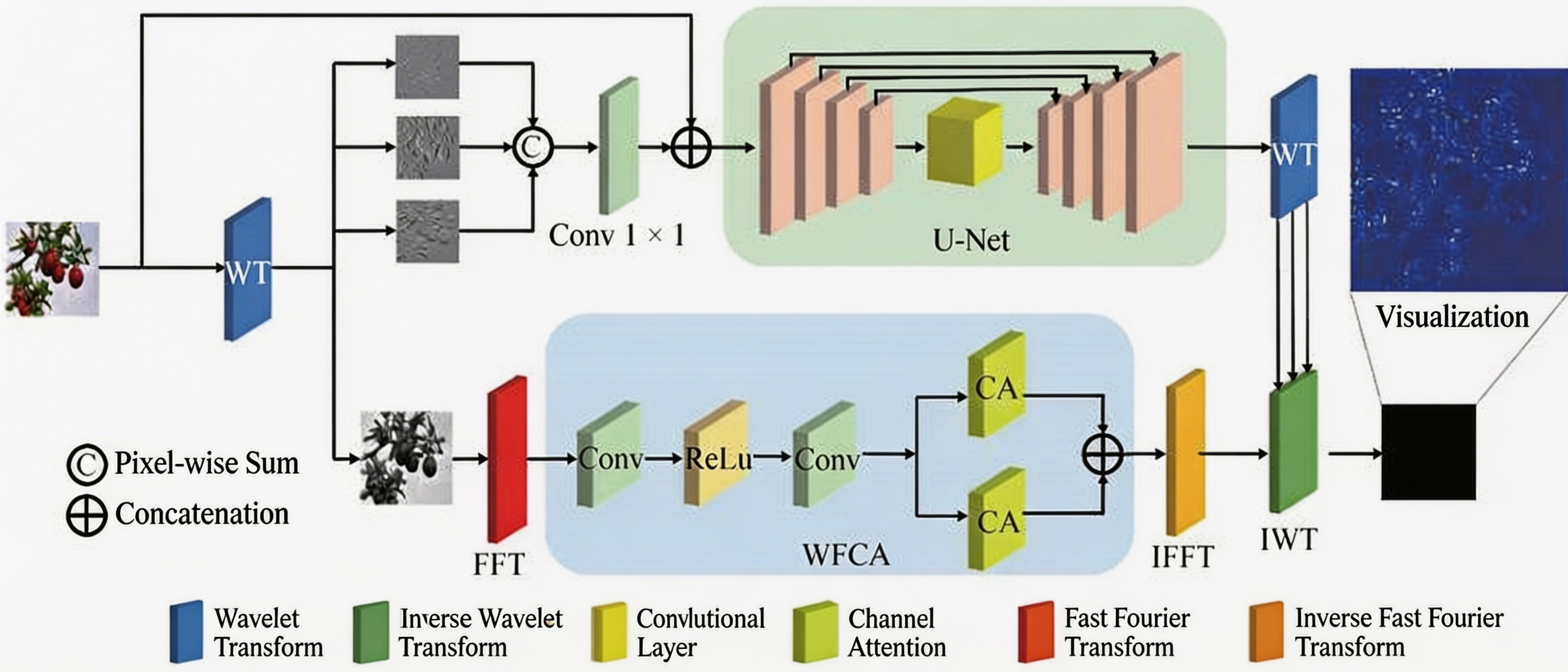} 
    \caption{Overall architecture of MSAAN}
    \label{fig:overall_arch}
\end{figure}

\section{Methodology}

\subsection{Overall Architecture}
The proposed Multi-scale Spatial Adaptive Attention Network (MSAAN) is designed to achieve high-quality image super-resolution while maintaining a lightweight structure. The overall framework, as depicted in Figure 1, comprises three primary components: the Shallow Feature Extraction Module (SFEM), the Deep Feature Extraction Module (DFEM), and the Image Reconstruction Module (IRM).

Given a low-resolution input image \(I_{LR} \in \mathbb{R}^{3 \times H \times W}\), the network first employs the SFEM, which consists of a single \(3\times3\) convolutional layer, to extract initial shallow features:
\begin{equation}
F_0 = H_{\text{SFEM}}(I_{LR}) \in \mathbb{R}^{C \times H \times W},
\end{equation}
where \(H_{\text{SFEM}}(\cdot)\) denotes the operation of the SFEM.

These shallow features are then fed into the DFEM, the core of MSAAN responsible for extracting rich hierarchical representations. The DFEM is constructed by stacking \(n\) Spatial Feature Mixers (SFMs):
\begin{equation}
F_i = H_{\text{SFM}_i}(F_{i-1}), \quad i = 1, 2, \dots, n,
\end{equation}
where \(H_{\text{SFM}_i}(\cdot)\) represents the function of the \(i\)-th SFM. A global residual connection is incorporated to facilitate the learning of high-frequency details and ease gradient flow, yielding the final deep feature:
\begin{equation}
F_{DF} = F_n + F_0.
\end{equation}

Finally, the IRM reconstructs the high-resolution output image from \(F_{DF}\). The IRM is implemented as a lightweight upsampling layer, containing a \(3\times3\) convolution followed by a PixelShuffle operation \cite{veturi2024ragbasedquestionansweringcontextual}. A skip connection from the bilinearly upsampled input image is added to aid convergence and enhance performance:
\begin{equation}
I_{SR} = H_{\text{Upsample}}(F_{DF}) + \text{bilinear}(I_{LR}),
\end{equation}
where \(H_{\text{Upsample}}(\cdot)\) is the upsampling module and \(\text{bilinear}(\cdot)\) denotes bilinear interpolation.

\subsection{Spatial Feature Mixer (SFM)}
The Spatial Feature Mixer (SFM) serves as the fundamental building block of the DFEM, designed to effectively integrate local details with global context. As shown in Figure 1, each SFM sequentially processes input features through three key sub-modules: a Local Enhancement Block (LEB), a Multi-scale Spatial Adaptive Attention Module (MSAA), and a Feature Interactive Gated Feed-Forward Module (FIGFF). For an input feature map \(X \in \mathbb{R}^{C \times H \times W}\), the output \(Y_{\text{SFM}}\) is computed as follows:
\begin{align}
    Z_1 = H_{\text{LEB}}(X), \\
    Z_2 = H_{\text{MSAA}}(\text{LN}(Z_1)) + Z_1, \\
    Y_{\text{SFM}} = H_{\text{FIGFF}}(\text{LN}(Z_2)) + Z_2,
\end{align}
where \(\text{LN}(\cdot)\) denotes Layer Normalization \cite{Rombach2022}. \(H_{\text{LEB}}(\cdot)\), \(H_{\text{MSAA}}(\cdot)\), and \(H_{\text{FIGFF}}(\cdot)\) represent the functions of the LEB, MSAA, and FIGFF modules, respectively.

\subsubsection{Local Enhancement Block (LEB)}
To strengthen the network's ability to model local geometric patterns without significant computational overhead, we propose a lightweight Local Enhancement Block (LEB). Inspired by the role of relative position embeddings in enhancing local modeling within vision transformers \cite{veturi2024ragbasedquestionansweringcontextual}, the LEB acts as an efficient form of positional encoding. It is implemented as a \(3\times3\) depthwise convolution with a residual connection:
\begin{equation}
Y_{\text{LEB}} = H_{\text{DWConv}_{3\times3}}(X) + X,
\end{equation}
where \(H_{\text{DWConv}_{3\times3}}(\cdot)\) denotes the depthwise convolution operation. This design augments local feature representation with minimal parameter addition.

\subsubsection{Multi-scale Spatial Adaptive Attention Module (MSAA)}
The Multi-scale Spatial Adaptive Attention Module (MSAA) is the core innovation for capturing and fusing features across multiple spatial scales. It consists of two cascaded components: the Global Feature Modulation Module (GFM) and the Multi-scale Feature Aggregation Module (MFA).
\begin{equation}
Y_{\text{MSAA}} = H_{\text{MFA}}(H_{\text{GFM}}(X)).
\end{equation}

\textbf{Global Feature Modulation Module (GFM):} The GFM aims to modulate input features to enhance the continuity and diversity of global texture representations. It employs a differential feature extraction strategy \cite{li2024real}. First, initial features \(X_1 = H_{\text{Conv}_{1\times1}}(X)\) are extracted via a \(1\times1\) convolution. A global context vector is obtained through Global Average Pooling (GAP). The difference between the local features \(X_1\) and this global context is computed, scaled by a learnable parameter \(\gamma\) (initialized to 0), and used for feature reweighting. The result is fused with the original features and activated by a GeLU function \cite{jiang2024longllmlinguaacceleratingenhancingllms}:
\begin{equation}
M_1 = H_{\text{GeLU}}\left(X_1 + \gamma \odot \left(X_1 - \text{GAP}(X_1)\right)\right).
\end{equation}
This process dynamically suppresses less informative interactions, allowing the network to focus on learning balanced global representations.

\textbf{Multi-scale Feature Aggregation Module (MFA):} The MFA is designed to adaptively aggregate features from local to global scales. To maintain efficiency, the input feature \(M_1\) is first split into four groups along the channel dimension: \([X_0, X_1, X_2, X_3] = H_{\text{Split}}(M_1)\).

Each group undergoes a scale-specific processing path. For the \(i\)-th group, an adaptive max pooling operation with stride \(2^i\) downsamples the feature to simulate a larger receptive field. A \(3\times3\) depthwise convolution extracts features at this scale, which are then upsampled back to the original resolution \(s\) via nearest-neighbor interpolation:
\begin{equation}
\widehat{X}_i = H_{\uparrow s}\left( H_{\text{DWConv}_{3\times3}\left( H_{\downarrow \frac{s}{2^i}}(X_i) \right) } \right), \quad i=0,1,2,3.
\end{equation}

The processed features from all four scales are concatenated, combining fine-grained local details with coarse-grained semantic information. A \(1\times1\) convolution fuses them:
\begin{equation}
\widehat{X} = H_{\text{Conv}_{1\times1}}\left( H_{\text{Concat}}([\widehat{X}_0, \widehat{X}_1, \widehat{X}_2, \widehat{X}_3]) \right).
\end{equation}

Subsequently, a spatially adaptive attention map is generated. Both the aggregated multi-scale feature \(\widehat{X}\) and a separately processed version of \(M_1\) (via a \(1\times1\) conv and GeLU) are transformed by GeLU activation. Their element-wise multiplication produces a channel-attentive map that highlights important features. A final \(1\times1\) convolution yields the output:
\begin{equation}
Y = H_{\text{Conv}_{1\times1}}\left( H_{\text{GeLU}}(\widehat{X}) \odot H_{\text{GeLU}}\left(H_{\text{Conv}_{1\times1}}(M_1)\right) \right),
\end{equation}
where \(\odot\) denotes element-wise multiplication. This enables the MSAA to holistically integrate multi-scale context, local details, and long-range dependencies.

\subsubsection{Feature Interactive Gated Feed-Forward Module (FIGFF)}
We redesign the standard transformer feed-forward network to be more efficient for image SR. The proposed Feature Interactive Gated Feed-Forward Module (FIGFF) incorporates shift convolution (Shift-Conv) \cite{10851487} and a feature gating (FG) mechanism.

The input \(X\) first passes through a Shift-Conv layer and a GeLU activation: \(Z = H_{\text{GeLU}}(H_{\text{Shift-Conv}}(X))\). The feature \(Z\) is then split into two channel groups: \([\widehat{Z}_1, \widehat{Z}_2] = H_{\text{Split}}(Z)\). One branch (\(\widehat{Z}_2\)) is refined by a \(3\times3\) depthwise convolution. The other branch (\(\widehat{Z}_1\)) interacts with the refined features via element-wise multiplication, promoting cross-feature information exchange:
\begin{equation}
\widehat{Z} = \widehat{Z}_1 \odot H_{\text{DWConv}_{3\times3}}(\widehat{Z}_2).
\end{equation}
A final Shift-Conv layer produces the output:
\begin{equation}
Y_{\text{FIGFF}} = H_{\text{Shift-Conv}}(\widehat{Z}).
\end{equation}
The FG mechanism reduces channel redundancy and selectively enhances critical features, improving local feature extraction with low computational cost.

\begin{figure}[htbp]
    \centering
    \includegraphics[width=0.8\textwidth]{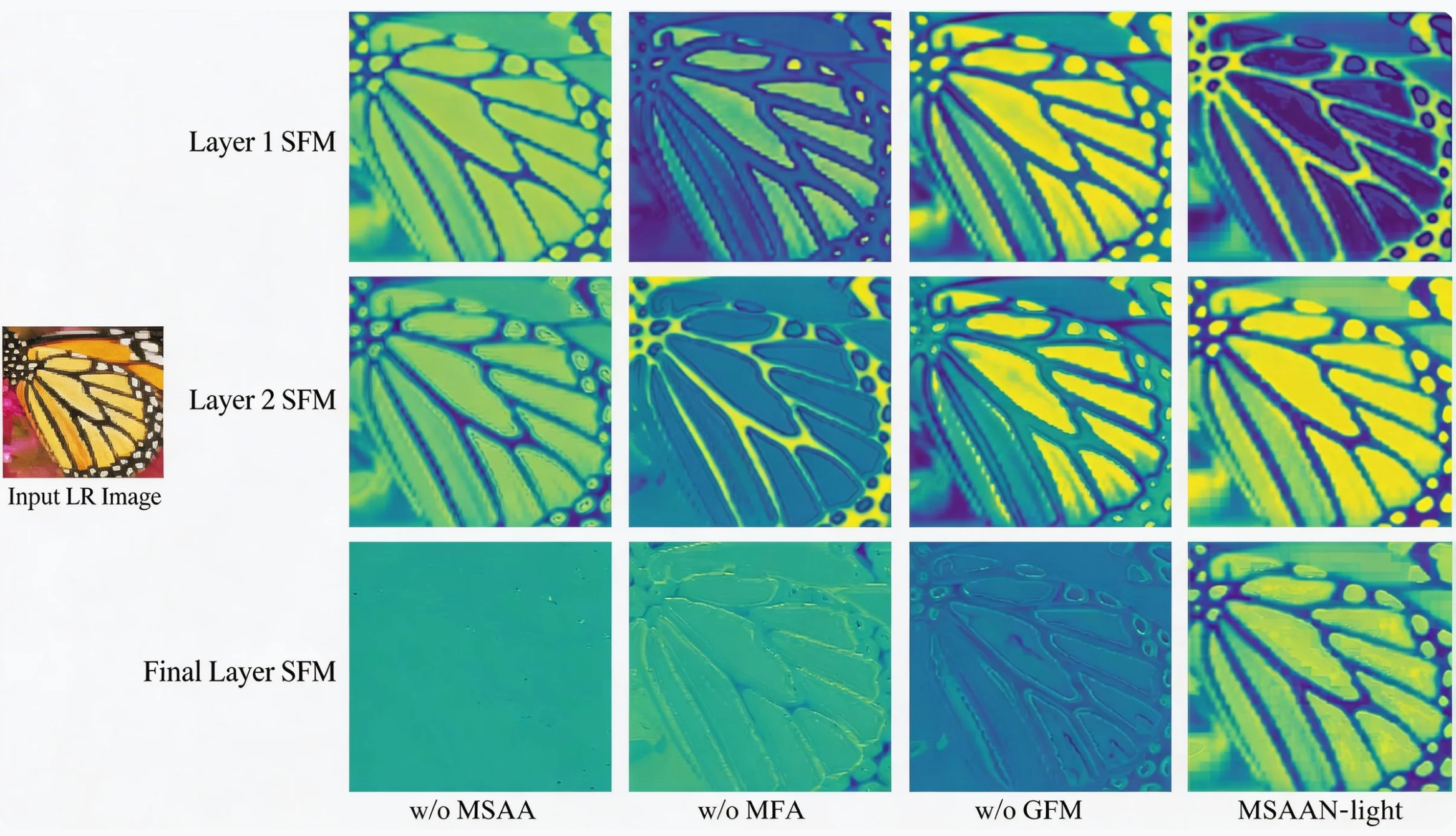}
    \caption{Visualization of feature maps before and after MSAA removal}
    \label{fig:feature_ablation}
\end{figure}

\section{Experiments and Results}

\subsection{Experimental Settings}

\textbf{Datasets and Metrics.}
The models are trained on the Flickr2K \cite{PyTorchTransforms} and DIV2K \cite{paddleocr2023} datasets, containing 800 and 2650 training images, respectively. Low-Resolution (LR) images are generated by applying bicubic downsampling to the High-Resolution (HR) references. For evaluation, we employ five standard benchmark datasets: Set5 \cite{jinensibieke2024goodllmsrelationextraction}, Set14 \cite{10.1145/3696271.3696299}, B100 \cite{zheng2023judging}, Urban100 \cite{huang2025mindev}, and Manga109 \cite{llama}. The reconstruction quality is assessed using Peak Signal-to-Noise Ratio (PSNR) and Structural Similarity Index (SSIM) calculated on the luminance (Y) channel after converting images to YCbCr color space. Model complexity is measured by the number of parameters and Floating-Point Operations (FLOPs).

\textbf{Model Configurations and Training Details.}
We train two versions of our network: a lightweight version \textit{MSAAN-light} (12 SFMs, 40 channels) and a standard version \textit{MSAAN} (24 SFMs, 60 channels). During training, random patches of size $64\times64$ and $48\times48$ are cropped from LR images for MSAAN-light and MSAAN, respectively. Standard data augmentation techniques, including random horizontal flipping and rotation, are applied. Both models are optimized using the Adam optimizer \cite{teknium2024hermes3technicalreport} ($\beta_1=0.9$, $\beta_2=0.99$). The initial learning rate is set to $1\times10^{-3}$ for MSAAN-light and $3\times10^{-4}$ for MSAAN, and is decayed following a cosine annealing schedule \cite{wu2024medicalgraphragsafe}. The loss function is a weighted combination of L1 loss and a Fast Fourier Transform (FFT)-based frequency loss \cite{10.1145/3655497.3655500}: $\mathcal{L} = \mathcal{L}_{1} + 0.05 \cdot \mathcal{L}_{FFT}$. All experiments are conducted using PyTorch on NVIDIA RTX 3090 GPUs.

\subsection{Ablation Studies}
Ablation studies are performed on the Set5 and Urban100 datasets with a scale factor of $\times4$, using the MSAAN-light configuration to validate the effectiveness of each proposed component.

\textbf{Number of SFMs.}
We investigate the impact of the number of stacked Spatial Feature Mixers (SFMs). As shown in Table 1, performance improves as the number of SFMs increases from 8 to 12. Beyond 12 SFMs, the performance gain becomes marginal or even slightly decreases, indicating that 12 SFMs represent a good balance between network depth and feature representation capacity for our lightweight design.

\begin{figure}[htbp]
    \centering
    \includegraphics[width=0.7\textwidth]{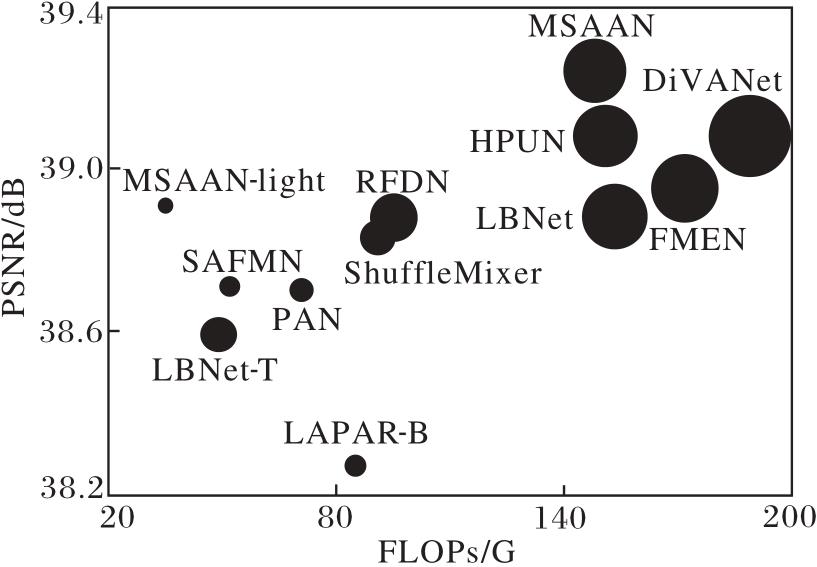}
    \caption{Comparison of metric values for different methods on the Manga109 dataset at a scale factor of ×2}
    \label{fig:scatter_metrics}
\end{figure}

\textbf{Effectiveness of LEB.}
The contribution of the Local Enhancement Block (LEB) is analyzed in Table 2. Adding the LEB improves PSNR by 0.04 dB and 0.06 dB on Set5 and Urban100, respectively, for MSAAN-light, with a negligible parameter increase of only 0.5K. A similar positive effect is observed when integrating LEB into SAFMN \cite{Sun2018}, confirming its general utility in enhancing local geometric modeling.

\begin{figure}[htbp]
    \centering
    \includegraphics[width=\textwidth]{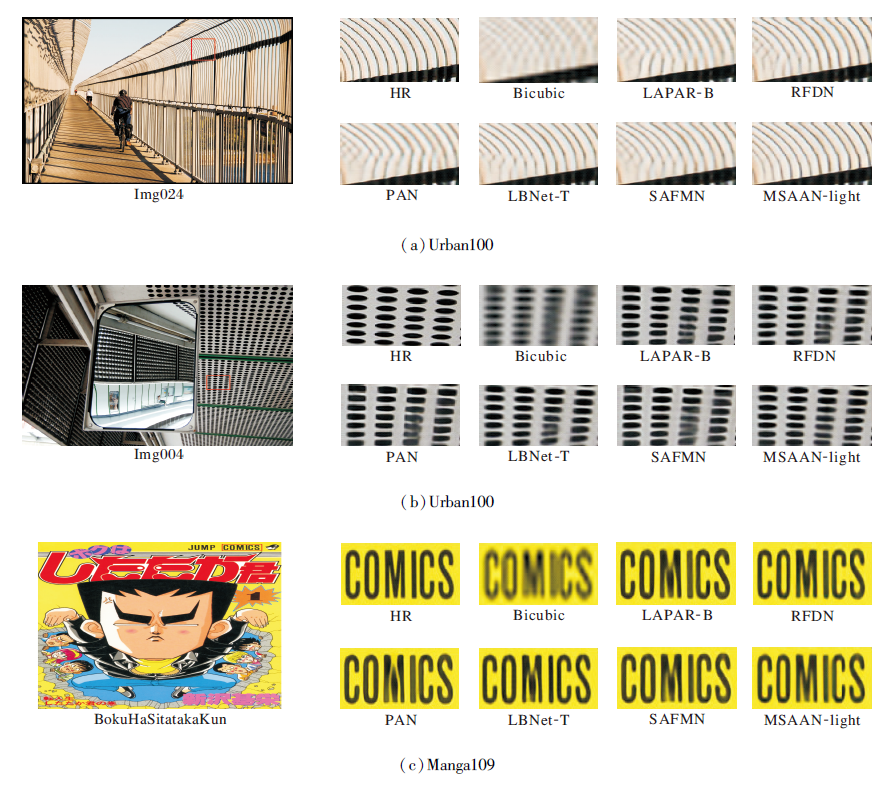} 
    \caption{Visualization results of MSAAN-light and contrastive methods at a magnification factor of 4}
    \label{fig:visual_lightweight}
\end{figure}

\textbf{Effectiveness of MSAA Components.}
We ablate the two core components of the Multi-scale Spatial Adaptive Attention Module (MSAA): the Global Feature Modulation Module (GFM) and the Multi-scale Feature Aggregation Module (MFA). Results in Table 3 demonstrate that removing either GFM or MFA leads to performance degradation. The complete MSAA module achieves the best results, verifying that GFM and MFA work synergistically; GFM provides well-modulated global context, while MFA enables effective multi-scale feature fusion.

\textbf{Effectiveness of FIGFF and FG Mechanism.}
Table 4 shows that replacing a standard MLP with our Feature Interactive Gated Feed-Forward Module (FIGFF) yields better performance. Furthermore, ablating the Feature Gating (FG) mechanism within FIGFF, as shown in Table 5, not only increases parameters and FLOPs but also degrades performance. This validates that the FG mechanism is crucial for reducing channel redundancy and facilitating beneficial feature interactions.

\subsection{Comparative Experiments}

\subsubsection{Quantitative Comparisons}
We compare MSAAN-light with several state-of-the-art lightweight SR methods: RFDN \cite{zhang2023blind}, LAPAR-B \cite{10692439}, PAN \cite{10871796}, ShuffleMixer \cite{10.1145/3696271.3696294}, LBNet-T \cite{10851615}, and SAFMN \cite{10851677}. The standard MSAAN is compared against larger models: LatticeNet \cite{10871796}, HPUN \cite{zhao2024harmonizing}, DiVANet \cite{tang2024mtvqa}, ESRT \cite{wang2025evaluating}, LBNet \cite{10823727}, FMEN \cite{feng2023unidoc}, and NGswin \cite{liu2025resolving}.

The PSNR/SSIM results for scale factors $\times2$, $\times3$, and $\times4$ are comprehensively listed in Tables 6 to 11. The key findings are summarized as follows:
\begin{itemize}
    \item \textbf{MSAAN-light vs. Lightweight Methods:} MSAAN-light consistently outperforms all competing lightweight methods across all scales and datasets while possessing fewer parameters and FLOPs. For instance, on the Manga109 dataset at $\times3$ scaling (Table 7), MSAAN-light surpasses RFDN, PAN, and ShuffleMixer by 0.13 dB, 0.19 dB, and 0.11 dB in PSNR, respectively, with 68\%, 33\%, and 58\% fewer parameters.
    \item \textbf{MSAAN vs. Larger Models:} As shown in Tables 9, 10, and 11, MSAAN achieves superior or highly competitive performance compared to models with comparable or even greater complexity. For example, on Manga109 at $\times3$ (Table 10), MSAAN outperforms ESRT, DiVANet, and NGswin by 0.28 dB, 0.29 dB, and 0.34 dB, respectively.
\end{itemize}
These results demonstrate that MSAAN achieves an excellent trade-off between reconstruction quality and model efficiency.

\begin{figure}[htbp]
    \centering
    \includegraphics[width=\textwidth]{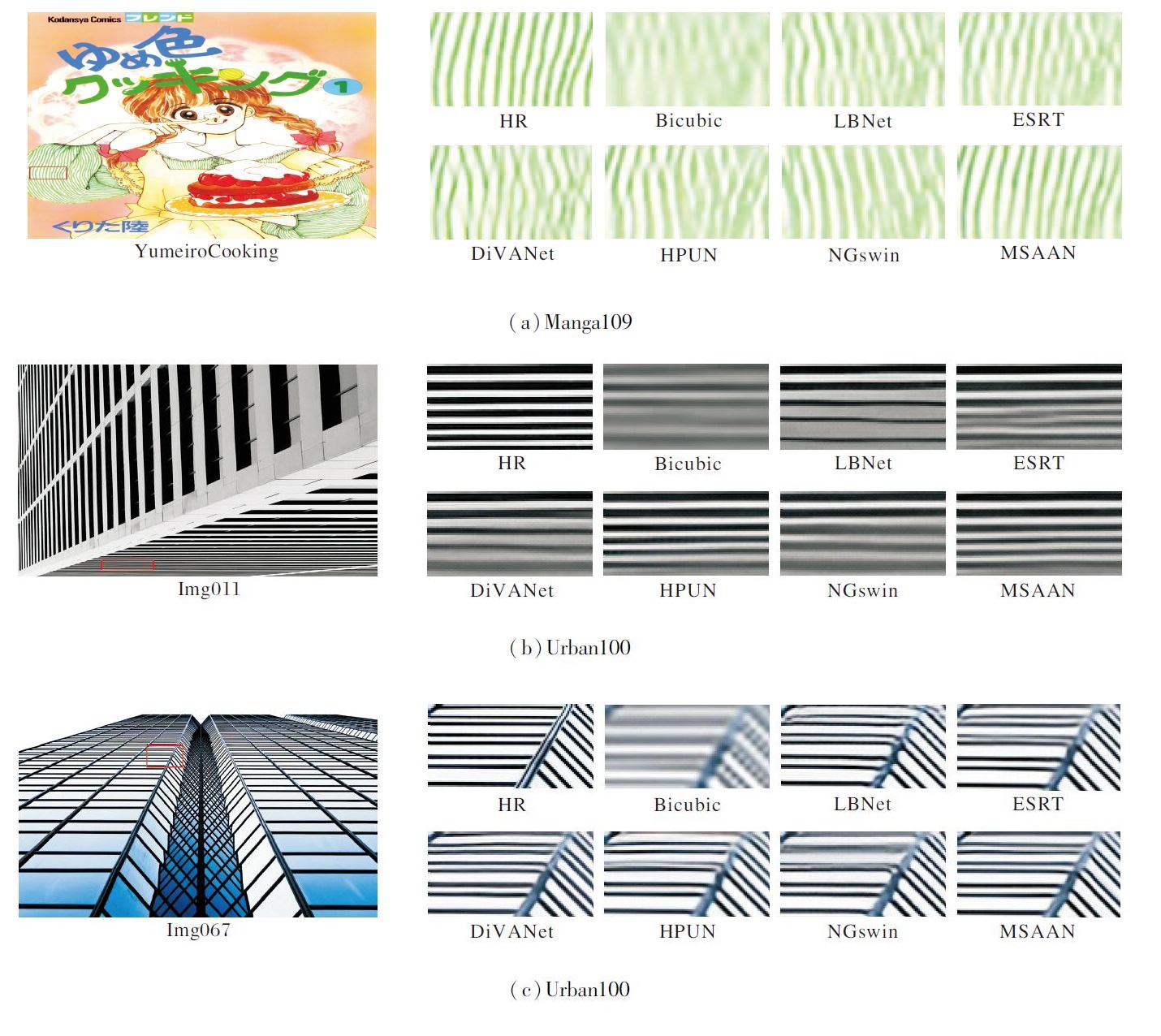} 
    \caption{Visualization results of MSAAN and contrastive methods at a magnification factor of 4}
    \label{fig:visual_large}
\end{figure}

\subsubsection{Visual Comparisons}
Visual results for $\times4$ SR are presented in Figure 4 (comparing MSAAN-light with other lightweight methods) and Figure 5 (comparing MSAAN with larger models). The proposed methods reconstruct images with sharper edges and more accurate textures. For example, in images containing regular patterns (e.g., stripes in `img\_024') or dense structures (e.g., `img\_004' and `img\_011'), MSAAN and MSAAN-light recover clearer and more consistent details compared to other methods, where results often appear blurry or contain artifacts.

\subsubsection{Analysis via Local Attribution Maps}
To understand which pixels contribute most to the reconstruction, we visualize Local Attribution Maps (LAM) \cite{AutoCAD2022} and compute the Diffusion Index (DI). As shown in Figure 6, the LAM for MSAAN-light covers a broader and more relevant region of the input image compared to RFDN, SAFMN, and PAN. The higher DI value for MSAAN-light indicates that our MSAA module enables the network to utilize contextual information from a wider pixel range for reconstruction, thanks to its effective integration of multi-scale and non-local features.

\begin{figure}[htbp]
    \centering
    \includegraphics[width=0.8\textwidth]{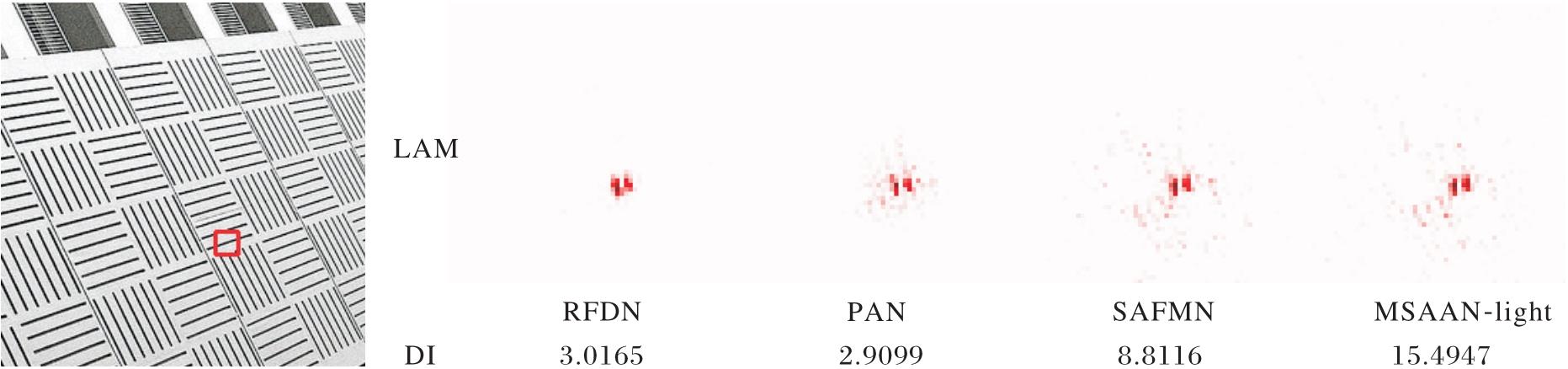} 
    \caption{Local Attribution Maps (LAM) of MSAAN-light and contrastive methods}
    \label{fig:lam}
\end{figure}

\subsection{Conclusion of Experiments}
Extensive experiments validate the effectiveness of the proposed MSAAN. Ablation studies confirm the contribution of each component (LEB, MSAA, FIGFF). Quantitative comparisons show that both MSAAN-light and MSAAN achieve state-of-the-art performance with high efficiency. Qualitative visual comparisons and LAM analysis further demonstrate the superior perceptual quality and more comprehensive use of image context by our method.

\section{Conclusion}

In this paper, we address the critical trade-off between reconstruction quality and model complexity in image super-resolution by proposing a novel Multi-scale Spatial Adaptive Attention Network (MSAAN). The core of our method is a carefully designed Multi-scale Spatial Adaptive Attention Module (MSAA), which effectively unifies the modeling of local high-frequency details and long-range contextual dependencies. The MSAA integrates a Global Feature Modulation Module (GFM) to learn coherent texture structures and a Multi-scale Feature Aggregation Module (MFA) to adaptively fuse features across scales. Furthermore, the auxiliary Local Enhancement Block (LEB) and Feature Interactive Gated Feed-Forward Module (FIGFF) are introduced to strengthen local geometric perception and enhance feature transformation efficiency, respectively.

Extensive experiments on multiple benchmark datasets demonstrate the superiority of the proposed MSAAN. Both the lightweight (MSAAN-light) and standard (MSAAN) versions achieve state-of-the-art or highly competitive performance in terms of PSNR and SSIM metrics while maintaining significantly lower parameter counts and computational costs compared to contemporary methods. Ablation studies validate the effectiveness of each proposed component. Visual comparisons and Local Attribution Map analysis further confirm that our network reconstructs images with sharper edges and more authentic textures by leveraging more comprehensive contextual information.

While the current model performs well on standard benchmarks, future work will focus on enhancing its generalization capability. This includes training with more diverse and realistic degradation models to better handle complex real-world scenarios. Exploring the integration of advanced attention mechanisms or dynamic network structures could also be promising directions for further improving performance and efficiency.

\clearpage

\nocite{*}
\bibliographystyle{IEEEtran}
\bibliography{custom}

\end{document}